\documentclass[10pt, a4paper]{article}
\usepackage{csquotes} 
\usepackage{tikzsymbols} 
\usepackage{arabtex} 
\usepackage{utf8} 
\setcode{utf8} 
\usepackage{lrec-coling2024} 

\title{ Cyber Risks of Machine Translation Critical Errors : Arabic Mental Health Tweets as a Case Study}

\name{Hadeel Saadany$^{\ast}$, Ashraf Tantawy, Constantin Or\u{a}san} 

\address{Birmingham City University, De Montfort University, University of Surrey \\
         United Kingdom \\
         $^{\ast}$hadeel.saadany@bcu.ac.uk\\}

\abstract{With the advent of Neural Machine Translation (NMT) systems, the  MT output has reached unprecedented accuracy levels which resulted in the ubiquity of MT tools on almost all online platforms with multilingual content. However, NMT systems, like other state-of-the-art AI generative systems, are prone to errors that are deemed machine hallucinations. The problem with NMT hallucinations is that they are remarkably \textit{fluent} hallucinations. Since they are trained to produce grammatically correct utterances, NMT systems are capable of producing mistranslations that are too fluent to be recognised by both users of the MT tool, as well as by automatic quality metrics that are used to gauge their performance. In this paper, we introduce an authentic dataset of machine translation critical errors to point to the ethical and safety issues involved in the common use of MT. The dataset comprises mistranslations of Arabic mental health  postings manually annotated with critical error types. We also show how the commonly used quality metrics do not penalise critical errors and highlight this as a critical issue that merits further attention from researchers.
 \\ \newline  }

\begin{document}

\maketitleabstract

\section{Introduction}

The accessibility of free machine translation (MT) systems such as Google Translate has profoundly changed how people engage with multilingual communication practices. One common use of the MT technology  is the translation of social media postings to overcome language barriers.  Another common use in research is the translation of social media data from a low-resource language  to a rich-resource language to utilise the more reliable tools trained for and developed by the rich-resource language  for the identification, extraction, analysis and labelling
of emotions and opinions \cite{arabicswn,arabictweetssentimentanalysis}.

However, there are substantial risks with this ubiquitous use of NMT systems that have not yet been sufficiently taken into account. Risks exist on two main levels. First, the state-of-the-art NMT engines are trained on huge amounts of data and have been shown to produce significantly \textit{fluent} mistranslations that are difficult to be realised by the end-user \cite{saadany2020great,al2022taxonomy,costa2022toxicity}. What exacerbates the problem is that users of MT online tools take the translations at face-value as there is no human intervention for accuracy checking.  Second, in safety-critical domains such as translation of mental health postings for detection of depression or suicidal ideation NMT critical errors can incur substantial damage by collectively leading  to false negatives where urgent mental-health problems pass undetected \cite{saadany2023analysing,delfani2024google}.

This research presents  our effort to bring awareness to the problem of critical translation errors, specifically in the translation of User-Generated Text (UGT) such as tweets concerned with mental health problems of the users. Our study addresses the MT critical errors on two levels. On the one hand, we analyse and extend an initial taxonomy specific of critical errors in the translation of tweets (Ref removed) with the objective of highlighting the cyber risks which the use of free online NMT tools may entail. On the other hand, we point to the fact that the commonly-used quality metrics can lend false confidence in the performance of  MT systems when their output includes critical errors. The aim of our study, therefore, is to shed light on the type of critical errors that can be produced in high-risk settings and to call for the case of introducing a metric that is capable of pointing out with some confidence measure to a critical error both on the user-level and on the MT  assessment level. We call on the research community to deploy our preliminary dataset and study to address the problem of  MT errors in safety-critical domains.

Thus, our contribution in this research is the following:

\begin{enumerate}
\itemsep0em
    \item Share an open-source dataset annotated for MT critical errors from the mental health domain \footnote{Reference to dataset 
anonymised}.
    \item Define and analyse critical errors  specific to the Arabic-English translation of UGT Twitter data.
    \item Present experiments of stress-testing several quality metrics on the critical error dataset and point to their shortcomings.
\end{enumerate}

To achieve its goals, the paper is divided as follows: first in section \ref{Critical} we define critical errors and summarise other research on MT critical errors. Then, in section \ref{sec:data} we explain how we compiled and annotated the critical error dataset. In section \ref{analysis} we explain the types of critical errors that were found in the dataset with examples. In section \ref{sec:QualityMetrics} we gauge the performance of the MT online tool with commonly used MT quality metrics and analyse each metric's performance with the different types of error in our compiled dataset. Finally in section \ref{sec:conclusion} we present our conclusions on the experiments and the intended future work to mitigate the risks of MT critical errors in the mental health domain.

\section{MT Critical Errors}
\label{Critical}

Recently, hallucinations by AI tools, such as ChatGPT, has been a part of a growing list of ethical concerns about AI. The same concern applies to the free MT engines frequently used on online platforms. Aside from misleading people with factually inaccurate information, MT hallucinations or critical errors can perpetuate biases or cause other harmful consequences if taken at face value. 

Recently, MT research started to address the problem of critical translation errors which can have grave consequences for users \citep{saadany2021challenges,sudoh2021translation,al2022taxonomy}. The realisation of the serious impact of this type of errors on users of NMT tools has even led to an organisation of a shared task for the prediction of sentences with catastrophic errors at the WMT 2021 \citep{specia2021findings}. One interesting finding  was the high proportion of critical errors in translation of UGT data \cite{specia2021findings}. Although this user-oriented line of research addresses different types of critical mistranslations, we focus in this study on the critical translation errors relevant to mental health postings on online platforms such as Twitter.

Tweets are one of the most popular datasets used for  monitoring users' psychological states for the detection and early prevention of suicide among both adolescents and adults. Research in this area has been predominantly in the English language \cite{desmet2013emotion,6918275,ji2020suicidal}. Due to conflicts  particularly in the Middle East and North African (MENA) region, social media platforms are increasingly used by the Arabic speaking population of this region to share thoughts about their mental health issues. Consequently, there have been an increasing interest in employing AI on Arabic UGT  to detect depression and suicidal ideation among the general public as well as refugees, migrants, and those who are forced to leave their homes due to massive humanitarian crises \cite{almouzini2019detecting,hassib2022aradepsu}.

In this research, we explore how far MT can help in the detection of Arabic tweet postings expressing depression and suicidal thoughts by their authors. We attempt to pin point cases where MT can lead to critical errors both for the user of an MT online tool and for a depression and suicide detection AI model trained on a corpus of tweets translated from Arabic to English. In this context, we define a machine translation critical error as:

\begin{displayquote}

\textit{An error that occurs when the MT engine output dangerously deviates from the intended message. Due to its fluency, the non-speaker of the source text cannot realise its existence and due to its consistent repetition in a corpus  it can skew an AI system to a higher percentage of false negatives when detecting depression or suicidal ideation. The resulting misinformation or misunderstandings have the potential to cause repercussions and lead to adverse public safety or health consequences.}

\end{displayquote}

In the next section, we explain how we compile a translated dataset of Arabic tweets and how we extract instances that fall within our definition of critical errors. We also show the linguistic cause of such errors which can be peculiar of the studied language pair.

\section{Data Compiling and Annotation}
\label{sec:data}

For error analysis, we use as our source data the AraDepSu dataset \cite{hassib2022aradepsu} which consists of 20K scraped tweets in different Arabic dialects as well as in Modern Standard Arabic (MSA). The dialects include Gulf, Egyptian, and Levantine. The dataset is compiled essentially for the detection of depression and suicidal ideation in Arabic tweets. The authors mention that the labels of the tweets are manually checked for correctness by human annotators and each record is labelled by one of the following categories: depression, depression with suicidal ideation, or non-depression. 

Since Google Translate (GT) is the most common MT application used  in real life as an open-source translator and since it is  used as the back-end of Twitter's MT tool, we opted for translating the dataset into English by GT. Then, in order to create a dataset for error analysis we followed a two-step semi-automatic approach. First, we build a depression and suicide detection classifier for English. Then, we use this classifier to predict the mental health label of the GT English translation of the Arabic tweets.  Instances where the gold label of the source tweet is “Depression with Suicidal Ideation” or ``depression" and the predicted label is “non-mental” were extracted as potential mistranslations by GT. The discrepant records are analysed manually to check the causes of the mistranslation if there are any. 

Thus, to narrow down the dataset for manual analysis, we fine-tuned an XML-Roberta-base model \cite{conneau2019unsupervised}  on a Reddit Mental Health Dataset \cite{low2020natural}. This dataset contains English posts from  mental health support groups on the Reddit blog. It was compiled to understand the impact of COVID-19 on mental health and to detect depression among these groups. Our training data consisted of $\approx$150k Reddit posts with the labels: `anxiety', `bipolar', `depression', `non-mental. This data was pre-processed by deletion of punctuation and non-alphanumeric symbols, lemmatisation, and lower-casing. We also transferred emojis to their equivalent lexicon ( (e.g. \dSadey[1.5] is translated into ``dislike”). We trained the model on four epochs and fine-tuned with the following AdamW  optimiser \cite{loshchilov2017decoupled} hyperparameters: learning rate = $1 \times e^{-5}$, $\beta1 = 0.9$, $\beta2 = 0.98$ and $\epsilon = 1 \times e^{-8}$.  We divided the dataset into 90\% training and 10\% validation set. The validation accuracy reached 92\%. The classifier’s predicted label was compared to the gold-standard label of the source text in the Arabic dataset. Records with discrepant labels between source and MT translation were extracted as either potential mistranslation or misclassification. From the discrepant instances,  $\approx$2200 tweets were annotated as mistranslations by our manual analysis and $\approx$800 from those mistranslations were annotated as critical errors as per our definition in Section \ref{Critical}. In the next section, we describe the analysis and the manual annotation of this dataset.

\section{Error Analysis}
\label{analysis}
To check causes of discrepancy in the extracted tweets and to pin down mistranslations that fall within our definition of critical errors, we hired an Arabic-English bilingual annotators to analyse the 2200 tweets. The annotation instructions were the following: 1) extract instances where the mistranslation of the Arabic source transfers a correct English sentence but dangerously deviates from the source message specially when the author is expressing a mental health issue, 2) based on our previous error typology for mistranslation of multilingual tweets (ref removed), assign the mistranslation with an error type, 3) correct the mistranslations by giving a correct English reference. 

\begin{figure}[t]
\begin{center}
\includegraphics[scale=0.6,trim={.15cm .4cm .1cm .73cm},clip]{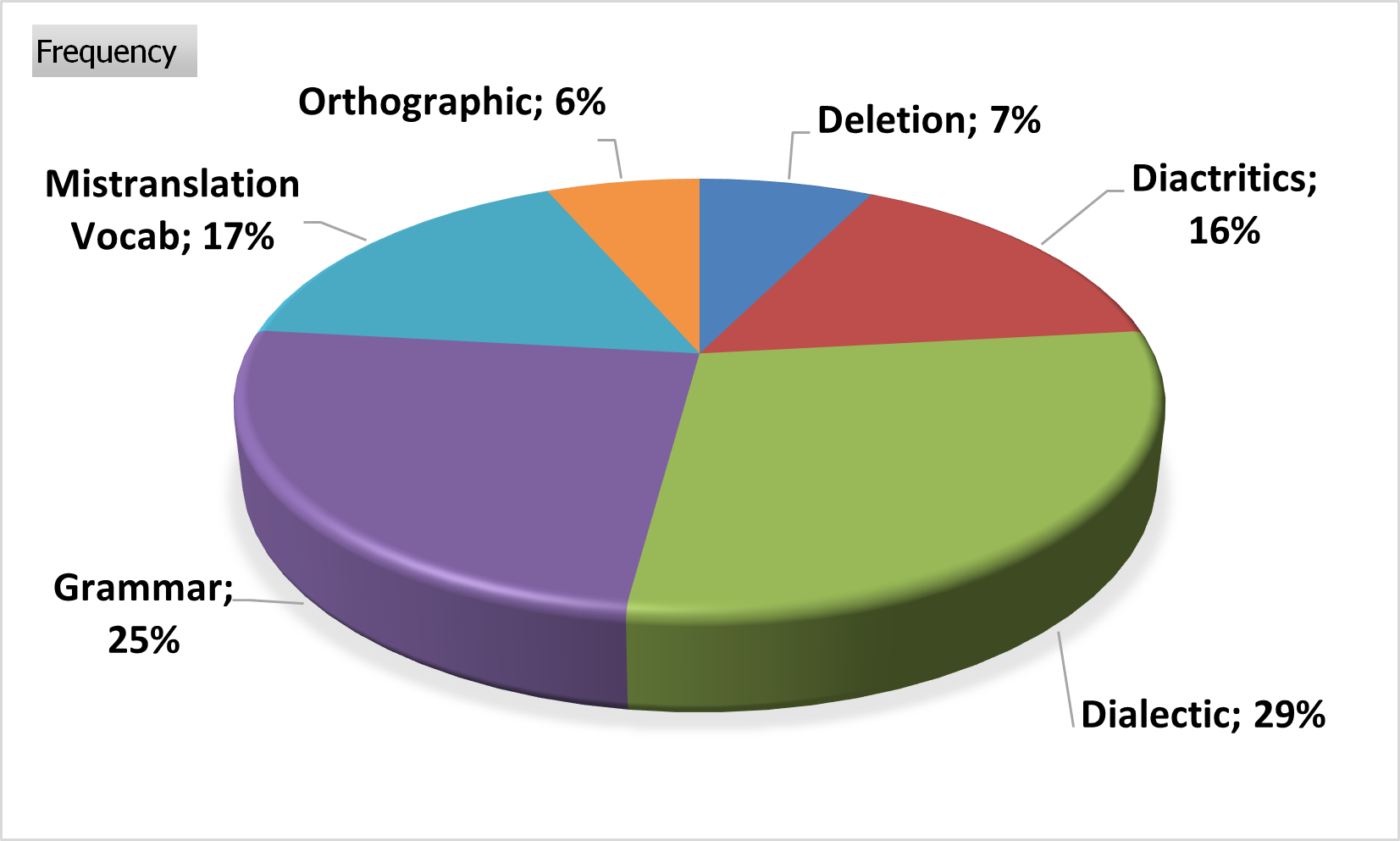} 
\caption{Frequency of Error Types}
\label{fig.1}
\end{center}
\end{figure}

Figure \ref{fig.1} shows the frequency of error types  as per the annotation schema we provided. In the next sections we explain each error category with examples\footnote{Due to space limitations, only one example is given for each error type. An appendix with several examples from the dataset can be included in the final version.}.
\subsection{Dialectical Language}
Around third of the critical errors in the dataset are due to the use of dialectical expressions. Research studies have shown that slang and dialectical expressions present several challenges to MT in general \cite{mtforarabicdialects}. Arabic Tweets are characterised by a wealth of slang expressions and code-switching between different dialects of one language based on the authors’ demographics. It was observed from the manual analysis that some dialectical expressions were consistently mistranslated. For example, the Gulf dialect of the verb ``want" is consistently translated as ``father". So tweets with clear suicidal inclinations such as `\<ما عندي حياه حلوه ابي انتحر >' and  `\<ابي الان اموت >', \    is translated as ``I don't have a sweet life, my father committed suicide"  and ``My father is now dead", respectively. The tweets should be translated as ``I don't have a good life, I want to commit suicide" and ``I want to die now". Clearly, the mistranslation obscures the alarming suicidal thoughts in the source. We hypothesis that this type of errors is due to the fact that the NMT system substitutes Out-of-Vocabulary (OOV)  with the highest contextual probability phrases. We consider these fluent mistranslations as critical errors both for users of the MT tool who would not be able to spot a wrong translation and for machine learning algorithms if the translated corpus is used for suicide monitoring and detection. 

Another high-risk error is the consistent mistranslation of the expression: `\<ابي اموت و افتك >' into ``I want to die and kill you". This should translate to ‘I want to die and be released’. The meaning is that the person wants to just escape the world and be released from it. However, GT consistently translates the word `\<افتك >'  (which means be released) as `kill you’. The critical error can have serious consequences to the authors of the tweets as they might be perceived as not only a danger to themselves but also a danger to others, which is not the case.

\subsection{Grammar Type Errors}
The mistranslations due to grammatical errors constituted 25\% and were found to fall within three  categories: mistranslation of negation, prepositions, or subject of the verb. The three types of errors lead to critical mistranslations, specially in tweets expressing suicidal intentions by their authors. For example, the Arabic tweet ``\< لا عايز انتحر بجد>" (I really want to  commit suicide) is translated into \mbox{``I really} don't want to commit suicide". The reason is that a negation marker ``\< لا>" (No) in the Arabic tweet is placed atypically  before the main verb. This is a special use in Arabic where the negative particle is used for assertion rather than negation \cite{jarrah2021favour}. In all instances where the negative particle is used as an assertive operator, the MT system provides a fluent English translation which transfers the exact opposite of the suicidal message in the source. Similarly, mistranslation of a preposition in tweets with `depression' label gives a wrong message. For example, the depression in the tweet `\<ابغي اعيش لحالي >' is lost in the wrong translation ``I want to live for myself" due to the swapping of the preposition `to' with `for'. The correct translation \mbox{``I want} to live by myself" should reflect the inclination to self-seclusion reflected in the source. As for the verb's subject, the mistranslation can cause a serious medical condition to go undetected. The tweet `\< بكمل يومين مو ماكله شي وفوق مو مشتهيه اكل >' (For two days, I didn't eat anything, I didn't crave food) reflects a possible depression symptom, i.e. loss of apetite. The GT mistranslation gives a syntactically correct sentence that completely misses the author's psychological state: `For two days, he didn't eat anything, and he didn't crave food'.

\subsection{Orthographic and Diacritic Errors}
The vowels in the Arabic language are realised by diacritics which indicate the pronunciation of the word. The same word can have different meanings based on the diacritic marks assigned, since a change in a diacritic is a change of a vowel sound. Arabic UGT is usually lacking diacritics since Arabic native speakers can easily guess which diacritic mark is intended based on the context of the word. The manual analysis has shown that the automatic translation often fails to realize the different meanings of words if diacritics are missing and this can leads to critical errors. For example, a tweet with `depression' label states `\<حتي الاكل مالي نفس اكل >' which means ``Even the food I don't have any appetite". The GT translates it as ``Even the food is not the same as eating", which is a syntactically correct sentence but completely deviates from the depressive mood expressed by the author. This is because the word `\< نفس>’ can either mean `appetite’ or `inclination to do something' if it has a `Kasraa’ (a short /i/) on the first letter or it can mean `the same’ if it has `fathaa’ (a short /a/ ) on the same letter.

We have similar critical errors due to the non-standard orthography of tweets. Linguists have observed that to encourage speed and immediacy of understanding, Twitter users type in the same way they speak \cite{theguardian_2010}. Thus, Arabic Twitter users not only eliminate diacritics but also write running sentences with no punctuation which again leads to critical mistranslations. For example, because of a lack of punctuation the tweet `\< ي اخي عاوز انتحر>' is translated as  ``My brother wants to commit suicide". The correct translation, however, is ``Oh Brother. I want to commit suicide". The elimination of full stops in the source text lead to a critical error as the translation gave misinformation that conceals the suicidal depressive mood of the tweet. The non-standard orthography of tweets leads to around 20\% of the errors in the dataset.

\subsection{Mistranslation of Vocabulary}

The analysis has shown that the mistranslation of vocabulary in short tweets, consisting of one or two words,  is the most critical since the distorted meaning is very difficult to recover. In such cases, the MT hallucinations are fluently incorrect, which makes the translation very dangerous in real-situations. For example, a common error in the dataset is translation of the one-word tweet `\< مشتت>' as ``dispersed"  when it actually means ``distracted" or ``cannot focus", a key symptom for some mental health issues. Similarly,  the  error of the two-word tweet `\<احس بضيق >' (I have anxiety) is equally critical as the GT  consistently translates it as ``I feel tight". Errors due to similar mistranslation of vocabulary items constituted 17
\subsection{Deletion Errors}
Research has shown that minimal deletion, even a single character deletion,  by the MT system can induce severe errors in the translation \cite{grissom2022rare}. Within the context of MT translation of mental health postings, deletion of a word or phrase can obscure a serious problem expressed by the source. For example, the MT output excludes the phrase `\<نفسي >' which means `I have a strong desire to do something' in the tweet `\<نفسي اقتل حالي >' ( I  wish to kill myself). The truncated translation `I kill myself' produced by the MT system misses the suicidal ideation in the source as it could simply mean that person is trying very hard to make things work. Critical errors due to deletion constituted only 7\% of the errors. From our analysis, we observed that with OOV, the MT system is more likely to insert a wrong phrase than to delete the unknown vocabulary all together.

\section{Quality Metrics Performance} \label{sec:QualityMetrics}
Recently, there has been an growing interest in assessing the ability of MT automatic quality metrics to detect critical errors in MT output \cite{saadany2021bleu,saadany2021sentiment}. The objective of this experiment is to measure the performance of existing quality metrics in detecting and penalising critical errors. Therefore, we stress-test some of the most commonly used quality metrics for machine translation on our dataset of critical errors. More specifically, we measure the performance of the following metrics: Sacrebleu, METEOR, Rouge, BERTScore, Google BLEU and TER. In the following, we briefly explain the metrics and their method of calculation, along with references for the interested reader. The following definitions for precision and recall for MT are used:

\textit{Precision}: The ratio of the number of matching n-grams to the number of total n-grams in the generated output sequence.

\textit{Recall}: The ratio of the number of matching n-grams to the number of total n-grams in the target (ground truth) sequence.

\paragraph{SacreBLEU} is the standard metric for assessing empirical improvement of MT systems \citep{papineni2002bleu}. It counts the n-gram matches between the machine and reference translations and the final score is a modified n-gram precision multiplied by a brevity penalty.

\paragraph{Meteor} is a multi-criteria matching metric which includes exact match, stem match, or synonymy match \cite{banarjee2005}. Its score is computed using an aggregation of uniform recall and precision, as well as a measure of fragmentation to capture the ordering of matched words. More recent versions (METEOR 1.5 and METEOR++2.0) apply also importance weighting by giving a smaller weight to function words \citep{denkowski2014meteor, guo2019meteor++}.

\paragraph{BERTScore} leverages the pre-trained contextual embeddings from BERT and matches words in candidate and reference sentences by cosine similarity \cite{bert-score}. We also calculate BERTScore Re-scaled which is re-scaled according to the contextual embeddings of the Common Crawl monolingual dataset\footnote{https://commoncrawl.org/}.

\paragraph{Rouge} is a recall-oriented metric as it compares the overlap of n-grams between reference and translation\cite{lin-2004-rouge}. We calculate ROUGE-L which takes into account sentence-level similarity and identifies longest co-occurring in sequence n-grams automatically.

\paragraph{GoogleBLEU} computes all sub-sequences of 1, 2, 3 or 4 tokens in output and target sequence with a sentence reward objective. The final score is  the minimum of recall and precision \cite{wu2016googles}. It has been shown to correlate well with the BLEU metric on a corpus level and performs better on the sentence level due to its sentence reward training objective.

\paragraph{TER} is the Translation Edit Rate which assesses the quality of the translation system in terms of the amount of editing that the translator would perform to change the MT output so that it exactly matches the reference translation \cite{TER}. 
Unlike all the  metrics mentioned above, with TER a higher score indicates more post-editing effort and hence a worse translation quality, so the lower the TER score the better.



\subsection{Results}

Table \ref{tab:my_label} shows the average score for each metric on the translated tweets with critical errors. The highest score is the BERTScore which gives an average of 0.91 to the translation, falsely indicating a very good performance of the MT system. The lowest score is the SacreBLEU which gives an average score of 0.30. However, given the fact that all the translations in the corpus have critical errors, the SacreBLEU score is still not indicative of the problems of the MT in this dataset. Overall, the average scores of all the studied metrics give a false confidence in an MT output with critical errors and hence these metrics may not be the optimal solution in assessing translation quality in high-risk uses of MT tools.

\begin{table}[tb]
    \centering
    \begin{tabular}{c|c}
        Metric & Average Score for all error types \\ \hline
        SacreBLEU  & 0.30 \\
        Meteor  & 0.56 \\
        RougeL & 0.59 \\
        BERTScore & \textbf{0.91} \\
        BERTScore\_sc & 0.51 \\
        TER & 0.55 \\
        Google\_BLEU & 0.34 \\ \hline
    \end{tabular}
    \caption{Average score for all error types for different metrics.}
    \label{tab:my_label}
\end{table}

\begin{figure*}[tb]
    \centering
    \includegraphics[scale=0.5, trim={0cm 0cm 0 0},clip]{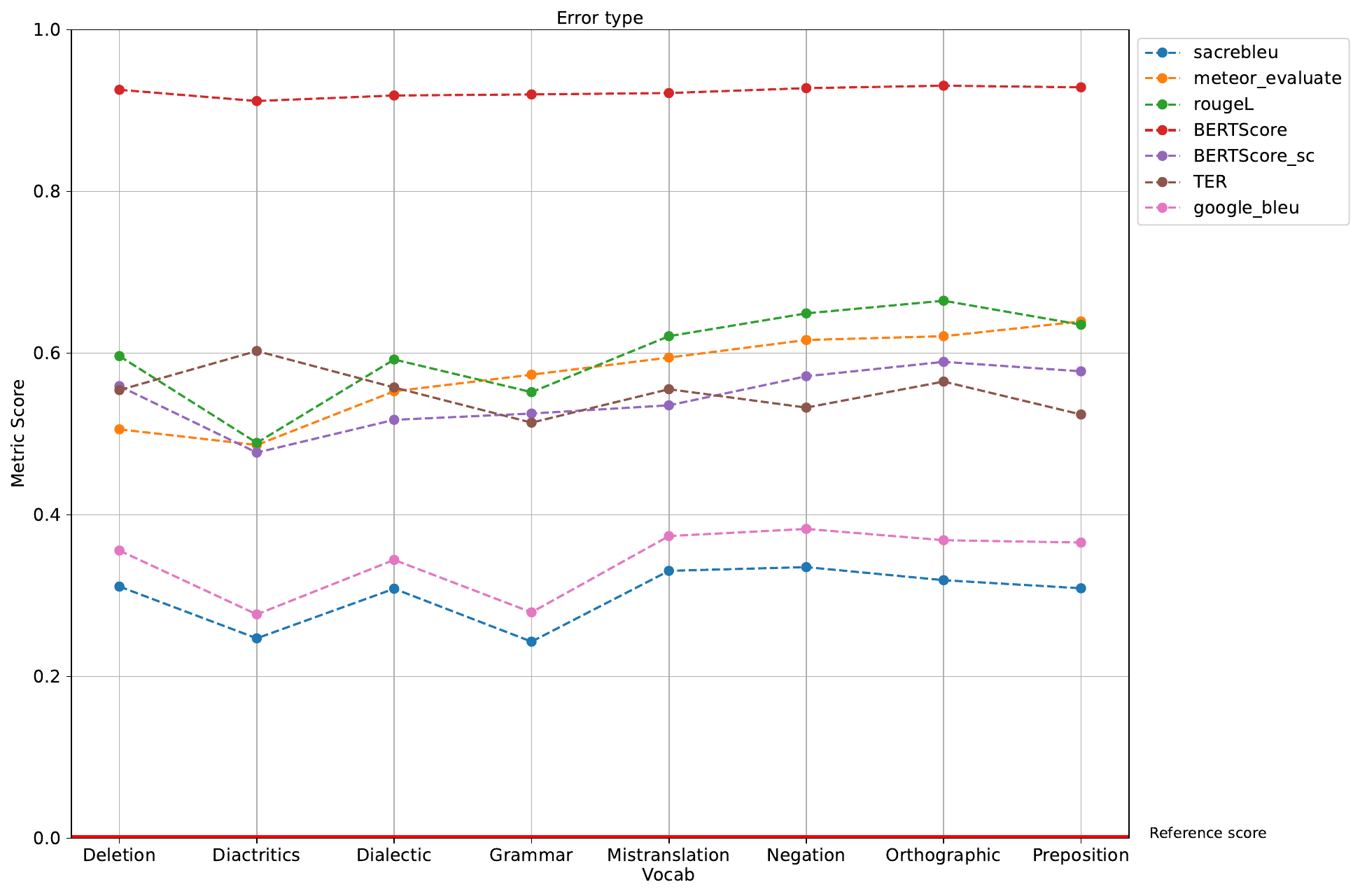}
    \caption{Metric scores for different types of errors}
    \label{fig:metricscore}
\end{figure*}

We also measured the average metric scores with respect to each error type we analysed in Section \ref{analysis}. Figure \ref{fig:metricscore} shows the performance of each metric versus different types of errors. With the exception of BERTScore, it can be noticed from the figure that there is a consistent pattern with regards to the different error types. Critical errors due to minimal change in the source text, namely n-gram deletion, change of a preposition, an orthographic marker or a negative particle, is the hardest to detect by the quality metrics since their scores are typically high with these errors. It is also worthy of notice that the MT critical errors with a higher edit distance, as indicated by the  higher score of TER with the diacritics type, causes a drop in  all the metric scores. Moreover, the SacreBLEU, the de facto metric for MT evaluation as well as the GoogleBLEU manifests better performance than the other quality metrics. Given the fact that all the assessed translations contain critical errors, the overall performance of the studied quality metrics highlights the need for a more adequate measure for the detection of critical errors in high-risk translations.






\section{Conclusion} \label{sec:conclusion}
Social media platforms are increasingly used across many population groups not only to communicate and consume information, but also to disclose psychological issues. There has been a growing interest in studying such data to monitor and detect mental health issues such as depression and suicidal ideation specifically in conflict-affected environments, such as the Arabic speaking regions of MENA. In this study, we address the problem of critical translation errors in Arabic tweets where the MT system produces a fluent mistranslation that deviates from the intended message. We specifically target such errors in high-risk settings where authors of the source message self-declare suicidal inclination or clear depression symptoms. We show from the analysis that  deviation from the intended message may spread misinformation, cause confusion and misunderstandings, and even possibly lead to detrimental consequences if the MT data is used for suicide and depression detection and prevention. 

We propose a taxonomy of critical errors and annotate a large amount of data with linguistic phenomena that can cause severe errors in the Arabic to English automatic translation of tweets. We also show that the commonly used translation quality metrics are incapable of giving a penalty proportional to the severity of errors caused by the mistranslation. We highlight this as a persistent issue among a wide range of MT evaluation methods which merits further attention from researchers. Thus, the  findings of our study prompt calls for action in three areas. First, we highlight a broad societal need for higher levels of awareness of the specific strengths and, crucially, of the limitations of MT. Second, we point to the potential consequences of uninformed MT use in  safety-critical domains such as healthcare specifically if the NMT result contains errors that are difficult to recognise. Third, our study calls for further research to improve the ability of evaluation metrics to detect and penalise critical errors both on the sentence level and on the corpus level.
Our future research will attempt to provide an evaluation mechanism that can detect critical MT errors where the mistranslation is too fluent to be recognised by MT users.

\nocite{*}
\section{Bibliographical References}\label{sec:reference}

\bibliographystyle{lrec-coling2024-natbib}
\bibliography{lrec-coling2024-example}

\bibliographystylelanguageresource{lrec-coling2024-natbib}

\end{document}